\documentclass{article}

\PassOptionsToPackage{numbers}{natbib}
\usepackage[preprint]{neurips_2024}

\usepackage[utf8]{inputenc} % allow utf-8 input
\usepackage[T1]{fontenc}    % use 8-bit T1 fonts
\usepackage{hyperref}       % hyperlinks
\usepackage{url}            % simple URL typesetting
\usepackage{booktabs}       % professional-quality tables
\usepackage{amsfonts}       % blackboard math symbols
\usepackage{nicefrac}       % compact symbols for 1/2, etc.
\usepackage{microtype}      % microtypography
\usepackage{xcolor}         % colors
\usepackage{amsmath}
\usepackage{tabularx}
\usepackage{threeparttable}
\usepackage{graphicx}
\usepackage[normalem]{ulem}
\usepackage{multirow}
\usepackage{wrapfig}
\usepackage{subcaption}
\usepackage{makecell}
\usepackage{enumitem}
\usepackage{bbm}
\usepackage{cleveref}
\usepackage{hyperref} 
\usepackage{wrapfig}
\usepackage{float}

\usepackage{caption}
\hypersetup{
  colorlinks   = true, %Colours links instead of ugly boxes
  urlcolor     = blue, %Colour for external hyperlinks
  linkcolor    = blue, %Colour of internal links
}
\usepackage{graphics}
\usepackage{graphicx}
\title{MathChat: Benchmarking Mathematical Reasoning and Instruction Following in Multi-Turn Interactions}

\author{Zhenwen Liang$^{1,2}$, Dian Yu$^{2}$, Wenhao Yu$^{2}$, Wenlin Yao$^{2}$, Zhihan Zhang$^{1,2}$, \\ \bf Xiangliang Zhang$^{1}$, Dong Yu$^{2}$\\
$^1$University of Notre Dame \,\,\,\, $^2$Tencent AI Lab, Bellevue \\
\texttt{\{zliang6,zzhang23,xzhang33\}@nd.edu} \\
\texttt{\{yudian,wenhaoyu,wenlinyao,dyu\}@global.tencent.com} \\
}
\newcommand{\mc}{\text{MathChat}}
\newcommand{\mcs}{\text{MathChat$_\text{sync}$}}

\begin{document}

\maketitle

% MathChat
% \mc
\begin{abstract}
Large language models (LLMs) have demonstrated impressive capabilities in mathematical problem-solving, particularly in single-turn question-answering formats. However, real-world scenarios often involve mathematical question-answering that requires multi-turn or interactive information exchanges, and the performance of LLMs on these tasks is still under-explored. This paper introduces \mc, a comprehensive benchmark specifically designed to evaluate LLMs across a broader spectrum of mathematical tasks. These tasks are structured to assess the models’ abilities in multi-turn interactions and open-ended generation. We evaluate the performance of various state-of-the-art LLMs on the \mc~benchmark, and we observe that while these models excel in single-turn question answering, they significantly underperform in more complex scenarios that require sustained reasoning and dialogue understanding. To address the above limitations of existing LLMs when faced with multi-turn and open-ended tasks, we develop \mcs, a synthetic dialogue-based math dataset for LLM fine-tuning, focusing on improving models' interaction and instruction-following capabilities in conversations. Experimental results emphasize the need for training LLMs with diverse, conversational instruction tuning datasets like \mcs. We believe this work outlines one promising direction for improving the multi-turn mathematical reasoning abilities of LLMs, thus pushing forward the development of LLMs that are more adept at interactive mathematical problem-solving and real-world applications.\footnote{Data and code are available at \url{https://github.com/Zhenwen-NLP/MathChat}. Work is done during Zhenwen's and Zhihan's internship at Tencent AI lab.}

% as well as developing AI systems that are not only proficient in mathematical problem-solving but also skilled in engaging in meaningful dialogues.

%  future math LLM enhancements
\end{abstract}

\section{Introduction}
Mathematical reasoning has been an essential task for computers for decades~\cite{boblow1968natural}. With the explosion in Large Language Model (LLM) development \cite{brown2020language,achiam2023gpt,touvron2023llama,touvron2023llama2,jiang2023mistral,team2024gemma}, mathematical reasoning has been widely recognized as a key ability for assessing these models. Most math reasoning benchmarks such as GSM8K \cite{cobbe2021training}, MATH \cite{hendrycks2021measuring}, SVAMP \cite{patel2021nlp}, MAWPS \cite{koncel2016mawps}, ASDiv \cite{miao2020diverse} and MathVista \cite{lu2023mathvista} feature the format of single-turn question answering (QA), where the input is a single question and the output is the solution. Recent studies~\cite{yu2023metamath,yue2023mammoth,gou2023tora,luo2023wizardmath,tang2024mathscale} have scaled up such QA data by distilling synthetic data from stronger LLMs like GPT-4 \cite{achiam2023gpt} or utilizing human-annotated datasets of rationales in diverse formats~\cite{yue2023mammoth,liang2023mint}, continually pushing the limits of math QA accuracy. For example, on one of the most widely recognized benchmarks, GSM8K, accuracy has increased from 10.4\% with a 175B-parameter model \cite{brown2020language} to 88.2\% achieved by a 7B-parameter model \cite{shao2024deepseekmath} in the past few years. 

%While these results seem promising, an important but under-explored question remains: 

% %\textit{\textbf{Can they reason and assist humans interactively within the mathematical domain? }}
%\textit{\textbf{What is the capability of LLMs on math reasoning in more-advanced settings?  }}

While math-specialized LLMs have shown promising progress on single-round QA benchmarks, their mathematical capabilities have not been verified in more complex scenarios. For instance, in real-world applications, such as interactive chatbots \cite{lee2022developing,janvcavrik2022artificial} and problem-solving assistants \cite{nguyen2019intelligent}, math tasks extend beyond single-round QA and require much more advanced reasoning and instruction following abilities such as dialogue understanding, diagnostic reasoning, educational feedback, etc. 
\emph{Can the established math-specialized LLMs perform as well on multi-round math reasoning as they do on single-round tasks?}
This question has not been comprehensively studied, although many recent studies have identified critical weaknesses of state-of-the-art LLM reasoners that could happen in multi-round interactions, such as long-context reasoning \cite{chen2024masked}, self-reflection ability \cite{huang2023large}, error identification \cite{openreview}, and educational content generation \cite{kasneci2023chatgpt}.

\begin{figure}[t]
    \centering
    % 第一个子图，左边
    \begin{minipage}[b]{0.35\textwidth}
        \centering
        \includegraphics[width=\textwidth]{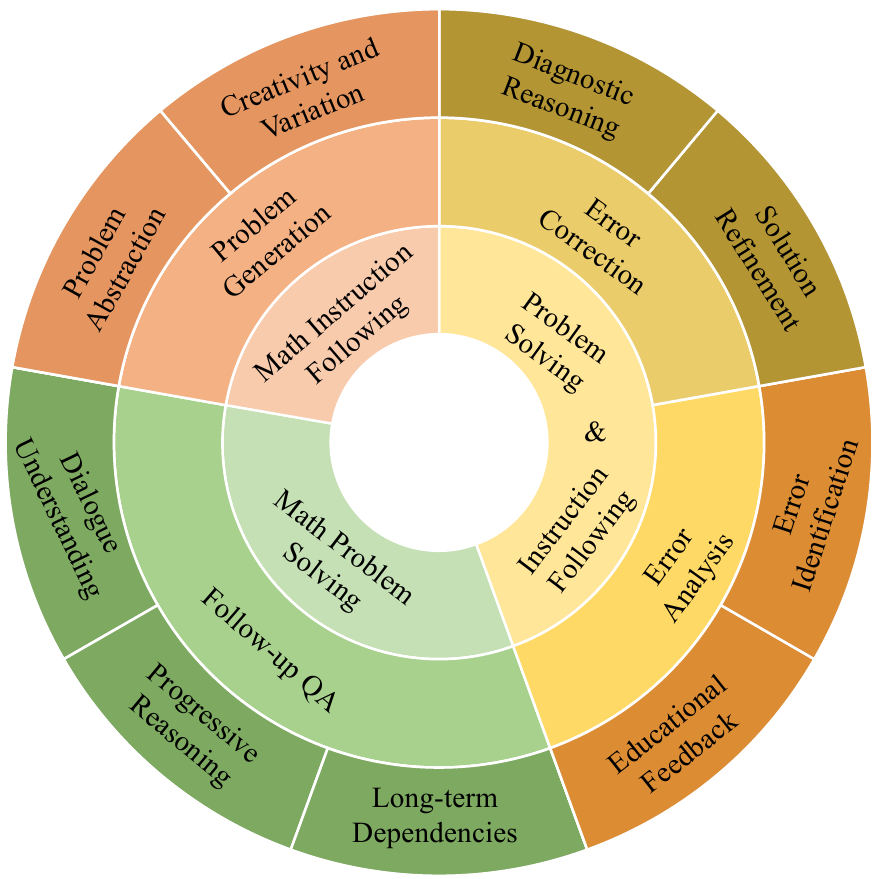}
        \vspace{-0.3cm}
        \caption{Taxonomy of \mc. The inner ring represents the task categories involved in \mc. The intermediate ring lists the evaluation tasks in \mc. The outer ring shows the tested capabilities in our tasks beyond simple math problem solving. See detailed descriptions in Section~\ref{dataset_overview}.}
        \label{fig:intro1}
    \end{minipage}
    \hfill
    % 第二个子图，右边
    \begin{minipage}[b]{0.6\textwidth}
        \centering
        \includegraphics[width=\textwidth]{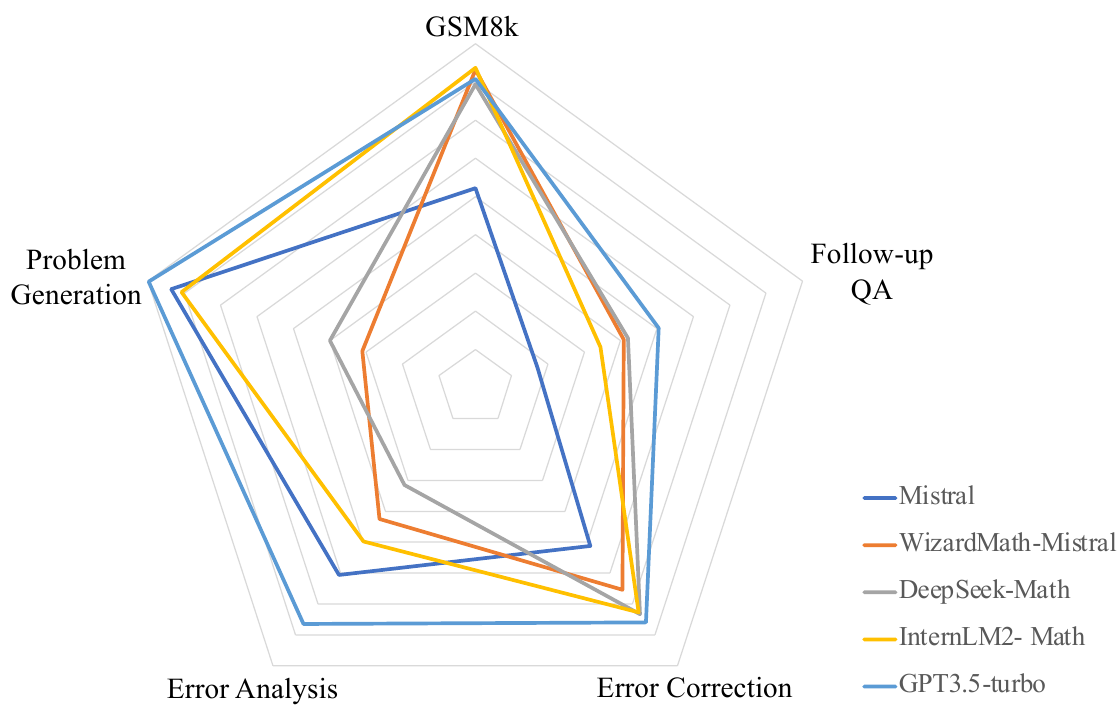}
        \vspace{-0.3cm}
        \caption{The performance comparison among various LLMs. Math LLMs (e.g., Deepseek-Math) have great performance on single-round QA dataset GSM8K, achieving similar performance to GPT-3.5. However, they significantly underperform GPT3.5 on \mc, which requires more advanced reasoning abilities. We average the evaluation metrics in each task and scale all values into 0-1 for better visibility.}
        \label{fig:intro2}
    \end{minipage}
    \vspace{-0.8cm}
\end{figure}

Therefore, in this paper, we advance the exploration of LLMs' mathematical reasoning abilities by introducing a new benchmark, \mc. Figure~\ref{fig:intro1} shows the hierarchical ability taxonomy derived from the tasks in \mc\ (e.g., those in Figure \ref{fig:example}), which are more advanced than the capabilities tested by single-round QA and addresses the above limitations noted in state-of-the-art LLMs.% regarding long-context reasoning, self-reflection, error identification, etc.

Based on our \mc~benchmark, we find that current state-of-the-art math-specialized LLMs that are fine-tuned on extensive mathematical QA data struggle to reliably answer multi-turn questions and understand instructions that extend beyond single-round QA. % though their performance on the single-turn QA dataset GSM8K is quite impressive, as shown in Figure \ref{fig:intro2}. 
Specifically, on open-ended tasks like \textsc{error analysis} and \textsc{problem generation} in  Figure \ref{fig:example}, the fine-tuned LLMs fail catastrophically since they can hardly understand the provided instructions. These shortcomings are perhaps unsurprising for models like MetaMath \cite{yu2023metamath}, which was trained exclusively on augmented question-answer pairs from single-turn math datasets GSM8K and MATH. The tasks in \mc~obviously represent a shift in distribution that challenges such models. However, even models like WizardMath \cite{luo2023wizardmath} that were trained on more diverse data including open-ended dialogues and evolving instructions fail to achieve satisfactory performance on \mc. We have also tried to reform our multi-turn math reasoning problem into a one-round math QA task by including all dialogue history in the question part, no significant performance improvement is observed. These results indicate potential over-tuning and data saturation towards the single-turn QA data inside current math LLMs, and also highlight a crucial open problem for the field of LLM development:

% GSM-style
\textit{\textbf{How can we empower math-focused LLMs to engage in multi-turn dialogues and follow diverse instructions without significantly compromising their problem-solving abilities?}}

% \textit{\textbf{How can we empower LLMs to engage multi-turn dialogue and diverse instruction following beyond just solving individual math problems?}}

To address the identified research challenge, we conduct an exploratory study to investigate various training data mixture strategies by leveraging extensive public math QA data, general-domain instruction tuning data, general-domain dialogue data, and our constructed synthetic dialogue-based math data (\mcs). The results 
%are presented in Figure \ref{fig:compare}, which 
indicate that the model trained with \mcs~significantly outperforms the baselines fine-tuned on other mixture datasets on open-ended tasks and surpasses the base LLMs on problem-solving tasks (see Section \ref{sec:sft} for more details). 

% our distinct supervised fine-tuning (SFT) data mixtures on LLMs: 1) Extensive public math QA datasets \cite{yu2023metamath,yue2023mammoth,mishra2022lila}. 2) Math QA datasets + general instruction tuning datasets \cite{peng2023instruction,zhou2023lima,ding2023enhancing}. 3) Math + Instruction Tuning + general dialogue datasets. 4) Math datasets in 1) plus a synthetic math problem-centered dialogue dataset created by us, which we named \mcs, generated by GPT models.

% \begin{wrapfigure}{r}{0.61\textwidth}
%   \centering
%   \includegraphics[width=0.61\textwidth]{./figures/compare.pdf}
%   \caption{Performance comparison of different Gemma-based models on \mc.}
%   \label{fig:compare}
% \end{wrapfigure}

In summary, this paper makes two main contributions. First, we introduce and release a benchmark \mc~dedicated to multi-turn math reasoning and conversation, aimed at advancing the development of a more generalized reasoner and assistant in mathematical contexts—a capability that existing math-specific LLMs currently lack. Second, we demonstrate that integrating synthetic math-dialogue dataset \mcs~with supervised fine-tuning (SFT) markedly enhances performance on open-ended tasks within \mc, without compromising much accuracy on direct problem-solving tasks. The resulting fine-tuned LLMs surpass their counterparts trained on various combinations of existing datasets. We believe this paper offers a new perspective on the evaluation of math-specific LLMs and advances the goal of developing a general math reasoning assistant.

\section{\mc}
%\subsection{Overview}
\label{dataset_overview}

We introduce \mc, designed to provide a deeper and more comprehensive examination of LLMs' abilities in multi-turn mathematical reasoning and instruction-following. \mc\ contains four novel tasks (\textsc{follow-up QA}, \textsc{error correction}, \textsc{error analysis}, and \textsc{problem generation}) inspired by previous studies in the education domain that reveal the importance of following a sequence of Initiate-Response-Follow-up~\cite{lim2020integral}, learning from self-made errors~\cite{heemsoth2016secondary}, and posing new problems with solutions~\cite{silver1994mathematical}. 
%These categories are designed to assess the advanced capabilities of LLMs, as shown in Figure \ref{fig:intro1}. 
The first two tasks focus on multi-turn mathematical problem-solving and reasoning, whereas the final two tasks evaluate the models' ability to follow mathematical instructions and respond to open-ended questions. All tasks within \mc~are sourced from the testing set of GSM8K, which we expanded using GPT-4 \footnote{We use gpt-4-0125-preview version in this paper.} to suit our specific requirements. All data in our benchmark is also provided with a reference answer from GPT-4 with post-verification. As a result, each task category contains the same number of samples as the GSM8K testing set—1,319. Table \ref{tab:stat} shows some basic statistics of our benchmark and Figure \ref{fig:example} shows some examples. All prompts used to generate the task data can be found in the Appendix \ref{app:task_generation}.

\begin{figure}
  \centering
   \vspace{-0.25cm}
   \includegraphics[width=0.95\textwidth]{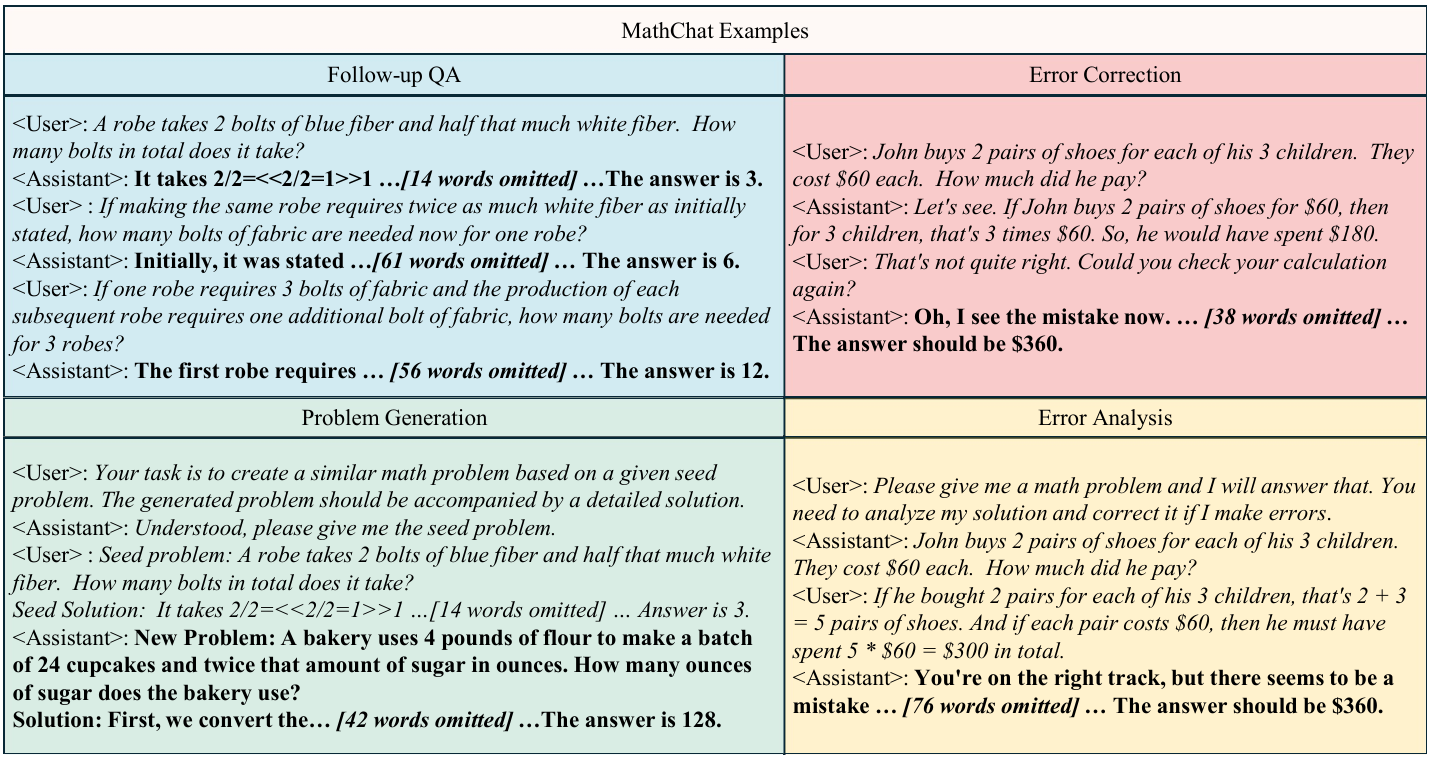}
  \caption{The examples of the four tasks in our \mc~benchmark. We place all tasks under a scenario of a dialogue between the user and the assistant. The part with a \textit{italics} font is the input to the LLMs and the \textbf{highlighted} parts are generated by LLMs and to be evaluated.}
  \vspace{-0.5cm}
  \label{fig:example}
\end{figure}

% -------------------- under construction -------------------- 
% Specifically, the first task focuses on \textsc{follow-up QA}. In this setup, we present a sequence of two in-depth follow-up questions beyond the initial question. The second task, \textsc{error correction}, is based on the scenario where the model initially provides an incorrect answer in the dialogue history. Then, the model is told that the previous answer is wrong and asked to give a corrected solution. The third task is \textsc{error analysis}. Beyond merely correcting an incorrect answer, the model is tasked with recognizing its mistake in the initial attempt, illustrating why it was mistaken, and then providing the correct solution. Lastly, we propose an open-ended \textsc{problem generation} task that requires the model to create a new mathematical problem and solution, closely related in theme or principle to a provided source exemplar. 
% -------------------- under construction -------------------- 

\paragraph{Follow-up QA}
In this task, we form a three-round dialogue between a human user and an AI assistant. The initial round consists of a question from the original GSM8K testing dataset, with its ground truth answer. We then use GPT-4 to generate two additional questions that require a deeper understanding of the original question, and they are added to the existing dialogue to initiate new rounds of conversations. The correct answers are produced by GPT-4 and subsequently verified and revised by two different LLMs (i.e., GPT-4 and Claude). After several rounds of verification and revision by the LLMs themselves, we successfully annotate $92.4\%$ of problems. For the remaining problems that do not meet the verification criteria, we manually create the follow-up questions and solutions to control the problem quality. After that, we randomly sample $50$ problems from this task and find that all of them are correct and high-quality. During the evaluation phase, we present only the questions to the models, and the answers generated by the evaluated LLM are used in the subsequent round as shown in Figure \ref{fig:example}, following methodologies similar to those used in general-domain benchmarks like MT-bench \cite{zheng2024judging}. 

\setlength{\intextsep}{0pt}
\begin{wraptable}{r}{0.488\textwidth}
\renewcommand\arraystretch{0.95}
\footnotesize
    \centering
    \caption{Average lengths in MathChat benchmark. The first-round QA is essentially GSM8k testing set. We can find that our MathChat has more informative answers than GSM8k.}
    \setlength\tabcolsep{3.7pt}
    \begin{tabular}{lc}
    \toprule
    Follow-up QA Question (First Round) & 46.25 \\
    Follow-up QA Question (Second Round) & 34.43 \\
    Follow-up QA Question (Third Round) & 41.60 \\
    Follow-up QA Answer (First Round) & 52.78 \\
    Follow-up QA Answer (Second Round) & 87.16 \\
    Follow-up QA Answer (Third Round) & 93.84 \\
    \midrule
    Error Correction Wrong Attempt & 54.82 \\
    Error Correction Mistake Correction & 75.27 \\
    \midrule
    Error Analysis Wrong Attempt & 66.17 \\
    Error Analysis Mistake Analysis & 94.69 \\
    \midrule
    Problem Generation New Problem & 55.37 \\
    Problem Generation New Answer & 105.13 \\
    \bottomrule
    \end{tabular}
    \label{tab:stat}
\end{wraptable}

% \setlength{\intextsep}{0pt}
% \begin{wraptable}{r}{0.495\textwidth}
% \small
%     \centering
%     \caption{Average lengths in MathChat benchmark. The first-round QA is essentially GSM8k testing set. We can find that our MathChat has more informative answers than GSM8k.}
%     \begin{tabular}{lc}
%     \toprule
%     Follow-up QA Question (First Round) & 46.25 \\
%     Follow-up QA Question (Second Round) & 34.43 \\
%     Follow-up QA Question (Third Round) & 41.60 \\
%     Follow-up QA Answer (First Round) & 52.78 \\
%     Follow-up QA Answer (Second Round) & 87.16 \\
%     Follow-up QA Answer (Third Round) & 93.84 \\
%     \midrule
%     Error Correction Wrong Attempt & 54.82 \\
%     Error Correction Mistake Correction & 75.27 \\
%     \midrule
%     Error Analysis Wrong Attempt & 66.17 \\
%     Error Analysis Mistake Analysis & 94.69 \\
%     \midrule
%     Problem Generation New Problem & 55.37 \\
%     Problem Generation New Answer & 105.13 \\
%     \bottomrule
%     \end{tabular}
%     \label{tab:stat}
% \end{wraptable}

% \begin{wrapfigure}{r}{0.4\textwidth}
%   \centering
%   \includegraphics[width=0.4\textwidth]{./figures/errortype.pdf}
%   \caption{The distribution of error types on error correction task.}
%   \label{fig:error}
% \end{wrapfigure}

\paragraph{Error Correction}
In this task, we present a question to GPT-4 and instruct it to intentionally generate an incorrect answer. The incorrectness of this response is easily verified by comparing it to the correct answer in the GSM8K dataset. We use this incorrect QA pair as the dialogue history and clearly state that the solution is incorrect, and then we prompt the AI assistant to correct the wrong answer. For this task, verifying the accuracy of the ground truth answer is unnecessary because the correct final response should match the original answer provided in the source dataset GSM8k. Thus, during evaluation, we simply check whether the LLM’s corrected answer aligns with the original answer. To ensure our benchmark contains a diverse set of error types, we conduct an analysis in Appendix \ref{app:error}.

\paragraph{Error Analysis}
%The initial QA pair for error analysis task is similar to that used in the error correction task, where the evaluated LLM is presented with an incorrect solution to a problem. However, the tasks diverge from the second round: while error correction solely focuses on rectifying the answer, error analysis requires the model to first recognize that an error exists (\textbf{\textit{Error Identification}}) and then analyze and correct it, providing \textbf{\textit{Educational Feedback}} on the nature of the mistake. Although these tasks may appear similar, they pose distinctly different challenges for LLMs, especially those specialized in mathematics. These models are predominantly trained to solve problems directly, aligning well with the goal of error correction. In contrast, error analysis demands that the model not only understand the instructions but also identify and articulate the nature of the mistake before correcting it. To enhance data diversity in our benchmark, we generate a different batch of incorrect attempts for the error analysis task using GPT-4, separate from those used in error correction.
LLMs have been proven to have weak error analysis abilities~\cite{huang2023large,openreview}.
The initial QA pair for the \textsc{error analysis} task is similar to that used in the \textsc{error correction} task, where the evaluated LLM is presented with an incorrect solution to a problem. However, the tasks diverge from the second round: while \textsc{error correction} focuses on rectifying the answer, \textsc{error analysis} further requires the model to first recognize that an error exists, then analyze the error and correct it. Although the two tasks share similarities in targeting errors, they pose distinctly different challenges for LLMs, especially those specialized in mathematics. These models are trained to solve problems directly, aligning well with the goal of \textsc{error correction}. In contrast, \textsc{error analysis} demands that the model not only understand the instructions but also identify and articulate the cause of errors before correcting them. To enhance data diversity in our benchmark, we generate a different batch of incorrect attempts for the \textsc{error analysis} task, separate from those used in \textsc{error correction}.

%\dian{analyze the overall error types for the two error-related subtasks; emphasize the diversity of errors (not merely the computation errors)}

% nature of the errors
\paragraph{Problem Generation}
The final task in \mc, Problem Generation, has been a direction of interest in both computer science and education for many years \cite{polozov2015personalized,wang2021math,zhou2023learning}. In this task, we provide the LLM with an original question-solution pair from the source dataset as part of the dialogue history. We then ask the LLM to create a new problem-solution pair that either delves deeper into the same topic or applies the same mathematical principles in a different context. %This task requires the LLMs to utilize \textbf{\textit{Problem Abstraction}} to understand and extrapolate the core principles from the original problem, and exhibit \textbf{\textit{Creativity and Variation}} in generating novel, relevant problems. 
This task is notably different from the typical mathematical QA, as it requires a model to generate questions rather than solve them. It challenges models to exhibit both creative and reasoning capabilities.

%With the four tasks described above, \mc~aims to benchmark LLMs not only as purely mathematical problem solvers but also as more effective assistants in mathematical reasoning and education. 

\section{Evaluation of Existing LLMs on \mc}
We assess a variety of baseline LLMs using the \mc~benchmark. Detailed experimental settings such as the descriptions of baseline models are located in Appendix \ref{app:baseline}.
\subsection{Evaluation Metrics}
For the problem-solving tasks (\textsc{Follow-up QA} and \textsc{Error Correction}), we extract the last numerical value that appeared in the model's response and compare it to the ground truth number. This approach aligns with the evaluation metrics used in most prior studies on math word problem solving. For the instruction-following tasks (\textsc{Error Analysis} and \textsc{Problem Generation}), we utilize GPT-4 to assign scores from $1$ to $5$ (higher is better) based on three carefully designed multi-dimensional criteria, similar to \cite{zheng2024judging,kim2024prometheus}. The \textsc{Error Analysis} task evaluates instruction following (IF), error diagnosis (ED), and solution accuracy (SA). The \textsc{Problem Generation} task assesses IF, SA, and problem quality (PQ).  A detailed description of these evaluation rubrics is available in Appendix \ref{app:task_eval}. All these metrics are measured on a scale of 1 to 5. Empirically, for instruction following tasks, a score of 1 to 2 indicates the failure to understand the instructions. A score of 2 to 3 signifies a basic understanding of the instructions, but the generated responses are often wrong. A score of 3 to 4 means the model has a good understanding of the instructions and can generate corresponding answers, though mistakes may still occur sometimes. A score higher than 4 indicates a very good response, which is usually fluent and relevant, with mistakes being rare.

\subsection{Prompting Template}
For math-specific LLMs like MetaMath and WizardMath, which are typically trained on specific QA templates, our \mc~involves multi-turn dialogues that do not strictly adhere to the formats of their training data. To fully exploit their potential in evaluation, we test these models in two settings: (i) using the chat template\footnote{\url{https://huggingface.co/docs/transformers/main/en/chat_templating}.} 
of their base models, and (ii) adapting their specific QA templates to include our dialogue history in the question part, i.e., reforming our multi-turn math reasoning problem to a one-round math QA task. For each task, we report  results from the better-performing setting. Empirically, we find that for tasks requiring problem-solving skills, such as \textsc{Follow-up QA} and \textsc{Error Correction}, the second setting significantly outperforms the first. However, performance is nearly identical across both settings for the instruction following tasks. These experimental evaluations reveal that solving the challenging tasks in  our benchmark  requires models to possess deeper understanding and comprehension abilities. For models that cannot perform well on our tasks, it is not merely due to their unfamiliarity with chat-template data. %Therefore, there is indeed intrinsic complexity and a demand for higher cognitive capabilities in solving tasks in our \mc.

\subsection{Result Analysis and Observations}

\begin{threeparttable}
\footnotesize
    \setlength\tabcolsep{6.4pt}
    \centering
    \caption{The performance of three open-sourced general-purpose LLMs, five math-specialized LLMs, and GPT-3.5-turbo on \mc. All open-sourced models are in the size of 7B. R1, R2, and R3 denote different rounds in Follow-up QA. Evaluation metrics: Acc. (\%), and others from 1 (lowest) to 5 (highest), such as IF = Instruction Following, ED = Error Diagnosis, SA = Solution Accuracy and PQ = Problem Quality. We \textbf{bold} the best performance achieved by open-sourced models. }
    \vspace{-0.2cm}
    \begin{tabular}{c c c c c c c c c c c}
        \toprule
        & \multicolumn{3}{c}{\makecell{Follow-up QA \\ R1\tnote{*} \;\;\;\;\; R2 \;\;\;\;\; R3 }} & \makecell{Error \\ Correction} & \multicolumn{3}{c}{Error Analysis} & \multicolumn{3}{c}{Problem Generation} \\
        \cmidrule(lr){2-4} \cmidrule(lr){5-5} \cmidrule(lr){6-8} \cmidrule(lr){9-11}
        & \multicolumn{3}{c}{Acc.} & Acc. & IF & ED & SA & IF & PQ & SA \\
        \midrule
        \multicolumn{11}{l}{\textit{General-Purpose 7B LLMs:}} \vspace{2pt} \\
        LLaMA2-chat & 15.09 & 11.67 & 8.12 & 38.82 & 2.64 & 1.83 & 1.87 & 4.02 & 3.83 & 3.33 \\
        Mistral-Instruct &32.06& 20.40 & 13.70 & 51.20 & \textbf{3.50} & \textbf{2.82} & 2.77 & \textbf{4.44} & 4.30 & \textbf{3.80} \\
        Gemma-it & 37.60 & 17.65 & 10.57 & 46.15 & 3.07 & 2.05 & \textbf{3.11} & 3.09 & 3.75 & 2.48 \\
        \midrule
        \multicolumn{11}{l}{\textit{Math-specialized 7B LLMs:}} \vspace{2pt} \\
        MAmmoTH & 66.85 & 32.16 & 19.31 & 54.15 & 2.55 & 1.75 & 1.79 & 2.03 & 1.95 & 2.42 \\
        MetaMath & 77.18 & 43.98 & 32.16 & 56.30 & 2.51 & 1.26 & 1.34 & 2.28 & 2.32 & 2.35 \\
        WizardMath& 83.20 & 44.81 & \textbf{36.86} & 68.22 & 2.62 & 1.81 & 1.95 &  1.53 & 1.54 & 1.60 \\
        DeepSeek-Math& 79.40 & \textbf{48.19} & 35.70 & \textbf{74.34} & 1.87 & 1.38 & 1.47 & 1.95 & 1.96 & 2.08 \\
        InternLM2-Math & \textbf{83.80} & 40.20 & 28.64 & 72.70 & 2.88 & 2.24 & 2.35 & 4.31 & \textbf{4.31} & 3.50 \\
        \midrule
        GPT-3.5-turbo & 74.68 & 55.26 & 45.59 & 75.90 & 4.12 & 3.64 & 3.71 & 4.62 & 4.62 & 4.23 \\ 
        GPT-4-turbo & 94.62 & 76.36 & 73.41 & 81.11 & 4.60 & 4.35 & 4.45 & 4.94 & 4.94 & 4.87 \\ 
        GPT-4o & 95.68 & 77.67 & 73.03 & 83.09 & 4.84 & 4.60 & 4.68 & 4.91 & 4.94 & 4.82 \\ 
        \bottomrule
    
    \end{tabular}
    
    \label{tab:baseline}
    \begin{tablenotes}
        \item[*] The first round performance is essentially the performance on the original GSM8K dataset.
    \end{tablenotes}
\end{threeparttable}

Overall, while most math-specific LLMs (except for MAmmoTH) outperform GPT-3.5-turbo only in the Round1 of Follow-up QA (see the first column in Table \ref{tab:baseline}), they fall short in all other tasks (other columns in Table \ref{tab:baseline}). These outcomes suggest that current math-specific models are overly tuned to single-round QA data, and the significant performance drop in multi-round and complex tasks further validates the rigor of our benchmark, challenging the models' diverse capabilities in mathematical reasoning, as illustrated in Figure \ref{fig:intro1}. We further investigate the models' performance across each task:

\paragraph{Follow-up QA.}
In Rounds 2 and 3 of the \textsc{Follow-up QA} tasks, models face significant challenges in multi-round math reasoning, with accuracy reductions ranging from 20\% to 50\%. This decline indicates that while math-specific LLMs initially outperform general-purpose LLMs and even GPT-3.5-turbo in Round 1, their performance deteriorates more significantly in subsequent rounds.
Theoretically, if a model maintains consistent accuracy across all dialogue rounds, with a first-round accuracy of $x_1$, the expected second-round accuracy would be $x_1^2$ due to error propagation. Interestingly, when comparing the square of the first-round accuracy ($x_1^2$) with the actual second-round accuracy ($x_2$), we observe a contrasting pattern: $x_1^2 > x_2$ for all math-specific LLMs, indicating a decline, whereas $x_1^2 < x_2$ for all other general-purpose models. This finding demonstrates that while math-specific LLMs excel at solving math problems in a single round, they show weaker progressive long-text reasoning capabilities within dialogues, highlighting a critical gap addressed by our benchmark.

\paragraph{Error Correction.}
In the \textsc{Error Correction} task, a clear distinction also exists between math-specific LLMs and general-purpose LLMs. Notably, general LLMs exhibit higher accuracy in correcting errors than in directly solving problems (i.e., the first-round follow-up QA), whereas the reverse is true for math-specific LLMs.
This adaptive behavior is evident in general-purpose LLMs but is noticeably lacking in math-specific LLMs, suggesting their weak ability to learn and reason from errors due to the over-tuning on single-round QA tasks. The difficulty of this task in our benchmark further emphasizes the need for models to go beyond single-round accuracy and develop robust error-correction abilities.

\paragraph{Error Analysis.}
The \textsc{Error Analysis} task requires that models first identify errors in a given solution before proceeding to analyze and correct them. In practice, we find that math-specialized LLMs often misinterpret the task's instruction about analyzing the solution and instead simply repeat the previous answer, or just validate the incorrect solution as correct. Conversely, only GPT-3.5-turbo relatively performs well in verifying the solution and pinpointing errors. This task presents a significant challenge for open-source mathematical LLMs, indicating a common limitation: their ability to identify and analyze errors. The high failure rate in this task also shows the challenging nature of our benchmark.

\paragraph{Problem Generation.}
The \textsc{Problem Generation} task, similar to \textsc{Error Analysis}, requires models to understand instructions that go beyond answering the given question. This task assesses several abilities: a model must accurately comprehend the given instruction, understand the provided problem-solution pair, and generate a new and relevant problem-solution pair. We observe that all general-purpose LLMs and only one math-specific model InternLM2-Math perform well. Other math LLMs, which are heavily optimized for problem-solving, struggle with this task. Empirically, we find that the above four models still consistently attempt to solve problems even when clearly instructed to create new problems. The difficulty of adapting to problem generation highlights the rigidity of current math-specific models, suggesting that these models are overly tuned to solve problems and, as a result, find it challenging to adapt to other tasks.
 %, which seems to be inherently simpler as it demands less complex reasoning and its reference-based nature.
% in almost all cases 
% 

%\end{document}

\section{Enhancement via Supervised Fine-Tuning}
\label{sec:sft}

Given the above challenges highlighted by our benchmark, it is natural to seek solutions to address these issues. In this section, we explore the performance improvements of general-purpose models enhanced by various supervised fine-tuning (SFT) strategies. See Appendix \ref{app:case} for case studies.

%To enhance the mathematical proficiency of math-specialized LLMs with comprehensive conversational capabilities, we leverage \mcs~to facilitate supervised fine-tuning strategies. 
 
\subsection{Baselines}
We first build a series of Mistral 7B baseline models  by applying  supervised fine-tuning with various existing datasets. First, \textbf{Mistral-Math} is developed to specialize Mistral-Instruct in math reasoning. This is achieved via fine-tuning the model by Arithmo \cite{arithmo} compilation, which includes three existing datasets: MetaMath \cite{yu2023metamath}, MathInstruct \cite{yue2023mammoth}, and Lila-OOD \cite{mishra2022lila}. The  used dataset totally comprises approximately 540,000 entries.
Second, \textbf{Mistral-Math-IT} is then built for enhancing the instruction following ability of Mistral-Math. We utilized the Alpaca-GPT4 dataset \cite{peng2023instruction}, which includes 52,000 instruction-following instances generated by GPT-4. We also use LIMA \cite{zhou2023lima}, which contains 1,000 high-quality prompts and responses from human interactions.
Last, \textbf{Mistral-Math-IT-Chat} gains the ability to engage in dialogue by tuning with chit-chatting dataset.
%We already discussed the limitations of math-specialized models in following diverse math-related instructions. 
%In this section, we further discuss how to equip LLMs with robust mathematical reasoning and instruction-following capabilities, as well as the ability to engage in dialogue. To achieve this, we adopt a straightforward approach by integrating three categories of datasets during the SFT stage. %We leave other resources and stages such as continued pretraining in future work.
%Consequently, we incorporate combinations of existing datasets specializing in math reasoning, instruction following, and chit-chatting as baselines. For mathematical reasoning, we selected the Arithmo \cite{arithmo} compilation, which includes three existing datasets: MetaMath \cite{yu2023metamath}, MathInstruct \cite{yue2023mammoth}, and Lila-OOD \cite{mishra2022lila}.
%The final math dataset used for SFT comprises approximately 540,000 entries. For general instruction following, we utilized the Alpaca-GPT4 dataset \cite{peng2023instruction}, which includes 52,000 instruction-following instances generated by GPT-4. We also use LIMA \cite{zhou2023lima}, which contains 1,000 high-quality prompts and responses from human interactions. 
%For the chat-style dataset, 
We subsample the Ultra-chat200k dataset \cite{ding2023enhancing} to 50,000 dialogues to minimize the training workload. Empirically, we find that this subsampling does not significantly affect performance on \mc~compared to using the entire Ultrachat-200k dataset.
Similarly, a series of Gemma 7B models are developed using the same SFT setting, and named following the same format. 

\subsection{Dialogue Dataset \mcs}
%From Table \ref{tab:sft}, we observe that SFT on the aforementioned datasets in mathematical reasoning, instruction following, and dialogue
%significantly enhances performance on problem-solving tasks such as \textsc{follow-up QA} and \textsc{error correction}. However, it 
%shows limited effectiveness in improving open-ended generation tasks , such as error analysis and problem generation. Consequently, we introduce and release a new dataset \mcs~specifically curated to address these challenges.
While the Ultra-chat200k dataset includes dialogues spanning a variety of topics, math-related conversations should be specifically highlighted and incorporated into the SFT process.  We thus introduce and release a new dataset \mcs, which is created by sampling QA pairs from Arithmo as seed examples. We then tasked GPT models to engage in real-world dialogues based on these seeds,  enriching the dataset with diverse and contextually relevant mathematical discussions. The details of the generation prompts are provided in the Appendix \ref{app:mathchat_generation}. Due to budget constraints, we generated 16,132 dialogues using GPT-4 and 131,346 dialogues using GPT-3.5-turbo, resulting in a total of 147,478 dialogues in the \mcs~ dataset. This dataset can serve as an augmented resource during the SFT stage
for future math LLMs, enabling them to engage in dialogues without compromising their ability to reason in single-round QA. Since \mcs~ already includes samples in forms of instruction and dialogue,  
Mistral and Gemma are tuned using both Arithmo and \mcs, resulting in \textbf{Mistral-MathChat} and \textbf{Gemma-MathChat} models, respectively.  

\subsection{Result Analysis and Observations}

\begin{threeparttable}
\footnotesize
    \setlength\tabcolsep{3.3pt}
    \centering
    \caption{Performance of LLMs that are fine-tuned with different datasets. The best performance is \textbf{bold} and the second best is \underline{underlined} for each series.
    }
    \vspace{-0.2cm}
    \begin{tabular}{c  c c c c c c c c c c | c}
        \toprule
        & \multicolumn{3}{c}{\makecell{Follow-up QA \\ R1\tnote{*} \;\;\;\;\; R2 \;\;\;\;\; R3 }} & \makecell{Error\\Correction} & \multicolumn{3}{c}{Error Analysis} & \multicolumn{3}{c}{\makecell{Problem\\Generation}} &\makecell{(Scaled)\\Overall}\\
        \cmidrule(lr){2-4} \cmidrule(lr){5-5} \cmidrule(lr){6-8} \cmidrule(lr){9-11} \cmidrule(lr){12-12}
        &  \multicolumn{3}{c}{Acc.}  & Acc. & IF & ED & SA & IF & PQ & SA & Average \\
        \midrule
        \multicolumn{12}{l}{\textit{Mistral 7B Series:}} \vspace{2pt} \\
        Mistral-Instruct &32.06& 20.40 & 13.70 & 51.20 & \textbf{3.50} & \underline{2.82} & \textbf{2.77} & \underline{4.44} & \underline{4.30} & 3.80 & 0.550\\
        Mistral-Math & 70.20 & 32.31 & 24.60 & \textbf{70.22} & 2.18 & 1.60 & 1.71 & 3.54 & 3.28 & 3.75 & 0.519\\
        Mistral-Math-IT & 70.73 & \underline{40.59} & \underline{27.74} & \underline{69.54} & 2.34 & 1.65 & 1.76 & 4.08 & 3.81 & 4.16 & 0.565\\
        Mistral-Math-IT-Chat & \textbf{71.79} & 39.22 & 27.36 & 69.15 & 2.31 & 1.50 & 1.63 & 4.39 & 4.20 & \underline{4.28} & \underline{0.574}\\
        Mistral-MathChat (Ours) & \underline{71.02} & \textbf{41.02} & \textbf{27.97} & 67.96 & \underline{3.40} & \textbf{2.89} & \underline{2.67} & \textbf{4.70} & \textbf{4.58} & \textbf{4.43} & \textbf{0.661}\\
        \midrule
        \multicolumn{12}{l}{\textit{Gemma 7B Series:}} \vspace{2pt} \\       
        Gemma-it & 37.60 & 17.65 & 10.57 & 46.15 & \underline{3.07} & \underline{2.05} & \textbf{3.11} & 3.09 & \textbf{3.75} & 2.48 & 0.463\\
        Gemma-Math & 70.73 & 29.70 & 19.92 & \underline{62.68} & 1.69 & 1.29 & 1.32 & 3.24 & 3.09 & 3.44 & 0.464\\
        Gemma-Math-IT & 72.02 & 43.36 & 32.57 & 62.60 & 1.76 & 1.40 & 1.46 & 3.34 & 3.32 & 3.61 & 0.508\\
        Gemma-Math-IT-Chat & \textbf{74.68} & \underline{46.35} & \textbf{33.64} & \textbf{63.85} & 2.05 & 1.64 & 1.70 & \underline{3.64} & 3.48 & \textbf{3.99} & \underline{0.549}\\
        Gemma-MathChat (Ours) & \underline{72.14} & \textbf{47.10} & \underline{32.64} & 61.86 & \textbf{3.43} & \textbf{2.90} & \underline{2.90} & \textbf{3.77} & \underline{3.72} & \underline{3.74} & \textbf{0.623}\\
        \bottomrule
    \end{tabular}
    \label{tab:sft}
    % \begin{tablenotes}
    %     \item[*] The first round is essentially GSM8K performance.
    % \end{tablenotes}
\end{threeparttable}

Table \ref{tab:sft} presents the results of two series of LLMs that have been fine-tuned from Mistral and Gemma models. The evaluation follows the same settings on \mc~as presented in Table \ref{tab:baseline}.
%LLMs, each fine-tuned with four distinct mixes of training data. 
Generally, the results suggest that our method of augmenting the training corpus enhances performance across all tasks. Notably, incorporating general-purpose instruction tuning data from sources such as Alpaca and UltraChat can improve performance on mathematical tasks. This improvement may stem partly from the inclusion of mathematical content within these datasets. The addition of high-quality instruction data predominantly may also boost the LLMs' natural language comprehension, thereby enhancing their ability to solve math problems. Moreover, the model fine-tuned with our \mcs~ dataset demonstrates markedly superior overall performance. Appendix \ref{app:overall} shows how we scale and calculate the overall score and Table \ref{tab:overall} contains a more comprehensive comparison in terms of the overall performance. Since \mcs~is created in a very simple and straightforward way, we believe that scaling up the quality and amount of such math dialogue data can bring more performance improvement, which we leave as our future work. 
Detailed analysis on each task follows.

\paragraph{Follow-up QA.}

When performing SFT with existing datasets, adding instruction-following, dialogue or our \mcs~datasets generally enhances the performance on follow-up QA tasks. Notably, we observe that performance improvements in the second and third rounds are significantly greater compared to the initial round of the original GSM8K QA. A likely explanation is that these datasets contain longer-context QA pairs, which enable the model to reason based on the dialogue history rather than focusing predominantly on more immediate contexts.

\paragraph{Error Correction.}

Fine-tuned models exhibit better accuracy than base LLMs in error correction, yet integrating additional datasets has not markedly boosted performance. This limited improvement suggests that essential skills such as \textsc{Diagnostic Reasoning} and \textsc{Solution Refinement}, indicated in Figure \ref{fig:intro1}, are not effectively learned from the used datasets.
Additionally, we observed that our \mcs~data negatively affects this task.
Upon examining the error cases, we discovered that models trained with \mcs~indeed have a better understanding of ``correcting the error'', where they try to make improvements over previous incorrect attempts rather than simply making new attempts. This contrasts with models trained purely on problem-solving datasets, which tend to give completely new solutions. 
The lower performance of the model trained with \mcs~may be attributed to the dataset's lack of manual filtering of incorrect cases. We leave the quality control problem and analysis in our future work.

\begin{figure}
    \centering
    \begin{minipage}[b]{0.95\textwidth}
        \centering
        \includegraphics[width=0.9\textwidth]{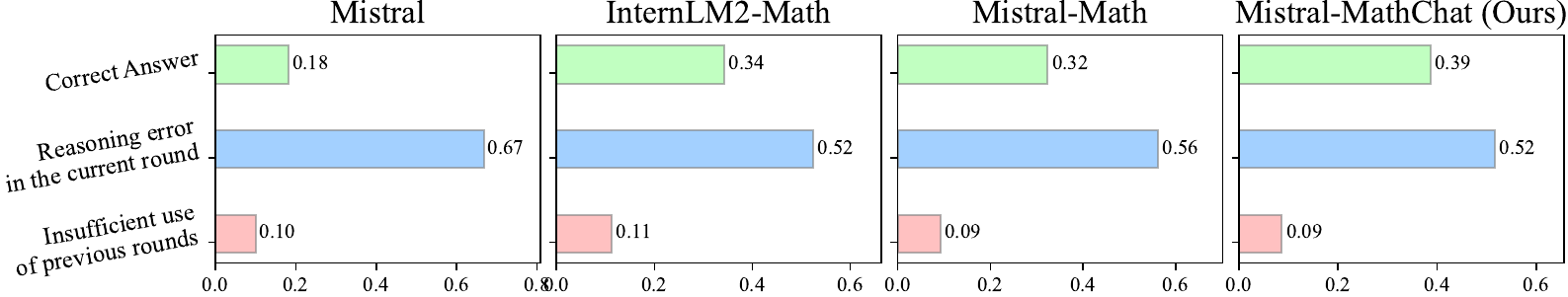}
        \caption{The Round3 answer quality in follow-up QA task.}
        \label{fig:output_1}
    \end{minipage}
    \par\vspace{0cm}
    \begin{minipage}[b]{0.95\textwidth}
        \centering
        \includegraphics[width=0.9\textwidth]{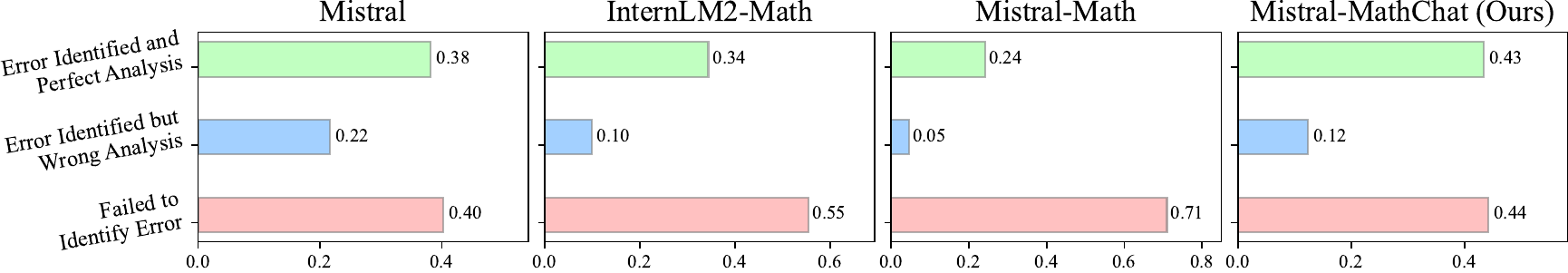}
        \caption{The answer quality in Error Analysis task.}
        \label{fig:output_2}
    \end{minipage}
    \par\vspace{0cm}
    \begin{minipage}[b]{0.95\textwidth}
        \centering
        \includegraphics[width=0.92\textwidth]{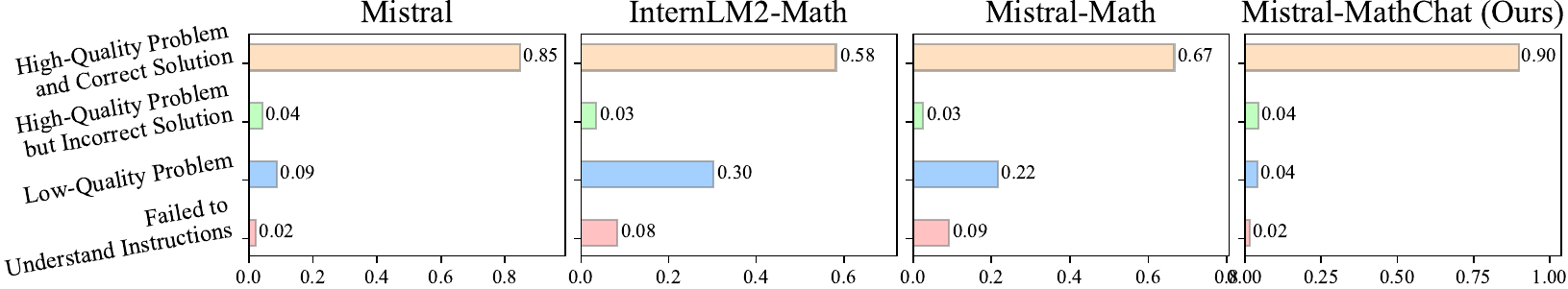}
        \caption{The answer quality in  Problem Generation task.}
        \vspace{-0.65cm}
        \label{fig:output_3}
    \end{minipage}
    %\caption{Combined output analysis for different tasks.}
    \label{fig:combined_output}
\end{figure}

\paragraph{Error Analysis.}
Similar to Error Correction, learning the ability to perform error analysis is challenging when using SFT with math QA and general instruction tuning datasets. Although the performance on this task is not exceptionally high, the inclusion of math-dialogue data in SFT has proven to be a viable method for enhancing LLMs' capabilities in error analysis. Our analysis in Figure \ref{fig:output_2} also reveals that the models fine-tuned with existing datasets (i.e., three baselines) typically affirm the correctness of previous answers and terminate their responses prematurely. In contrast, our \mcs~dataset aids LLMs in understanding how to conduct error analysis.

\paragraph{Problem Generation.}
On problem generation task, we observe that the base models already have reasonable performance and SFT without \mcs~generally hurts the performance. However, a particularly notable finding is the increase in Solution Accuracy (SA) scores following SFT, which suggests that fine-tuning on mathematical data helps the model recognize the importance of solution correctness and extend this awareness to generation tasks. Furthermore, our MathChat-enhanced SFT model records the best performance on this task, demonstrating the versatile utility of dialogue-enhanced training in mathematical contexts.

\vspace{-0.15cm}
\section{Analysis of Answer Qualities}
\vspace{-0.1cm}
To evaluate the answer qualities of various models on our MathChat benchmark, we analyzed 500 outputs each from Mistral, InternLM2-Math (i.e., the best math-specialized LLM in Table \ref{tab:baseline}), Mistral-Math, and our Mistral-\mcs~ model across tasks such as Follow-up QA, Error Analysis, and Problem Generation. We employed GPT-4 to categorize these outputs according to a predefined set of output categories. Our analysis revealed that the Mistral-\mcs~ models excel in tasks requiring open-ended responses, like error analysis and problem generation, while performing comparably in problem-solving tasks. The following sections detail these results:

\textbf{LLMs + \mcs~SFT achieves state-of-the-art accuracy in follow-up QA.} As shown in Figure \ref{fig:output_1}, all three math-specific models significantly outperform the original Mistral model, with our \mcs~model slightly surpassing the other two, showing the strong mathematical problem solving ability is still maintained after \mcs~fine-tuning.

\textbf{LLMs + \mcs~SFT exhibits strong error identification and correction abilities.} Figure \ref{fig:output_2} shows that although the Mistral model identifies errors in mathematical problems, it falls short in offering corrections. InternLM2-Math and Math-SFT show reduced error detection capabilities due to their intensive training on straightforward math QA. In contrast, our \mcs~model demonstrates a robust capacity for both identifying and correcting errors.%, achieving the highest completion rate in error analysis tasks.

\textbf{LLMs + \mcs~SFT demonstrates superior performance in problem generation.} As shown in \ref{fig:output_3}, our \mcs~model excels in problem generation tasks, while the other two math-specific models (InternLM2-Math and Math-SFT) struggle with instruction following and basic comprehension, highlighting the effectiveness of our \mcs~fine-tuning approach.

\vspace{-0.15cm}
\section{Related Work}
\vspace{-0.1cm}
\paragraph{Mathematical Reasoning.}
After the emergence of deep learning, Seq2Seq models \cite{wang2017deep,xie2019goal,alghamdi2022armath,liang-etal-2022-mwp,liang2022analogical,aaai,liang2023unimath} becomes popular in mathematical reasoning. Then LLMs have demonstrated success in solving math word problems through techniques like Chain of Thought (CoT)~\cite{wei2022chain,kojima2022large}, Program of Thought (PoT) \cite{chen2023program}, and sampling methods \cite{wang2022self}. These studies primarily focus on improving performance via better prompting design or inference strategies. Another line of research explores distilling synthetic data from LLMs to train smaller models \cite{magister2022teaching, liang2023let,luo2023wizardmath,yue2023mammoth}.%stronger LLMs are leveraged to generate high-quality instructions and reasoning steps (e.g., rephrasing problems, integrating tools), which are further used to supervised fine-tune smaller models. 
Some researchers also attempted extensive pre-training on math-related corpora to obtain foundational mathematical LLMs \cite{lewkowycz2022solving,taylor2022galactica,azerbayev2023llemma}.
As for the evaluation of mathematical reasoning, popular benchmarks include GSM8K, MAWPS \cite{koncel2016mawps}, MATH \cite{hendrycks2021measuring}, SVAMP \cite{patel2021nlp}, MathVista \cite{lu2023mathvista}, MathVerse \cite{zhang2024mathverse}, etc., and all of them are in single-round QA format. State-of-the-art (SOTA) models such as MetaMath \cite{yu2023metamath}, WizardMath, MathInstruct \cite{yue2023mammoth}, ToRA \cite{gou2023tora}, OpenMathInstruct \cite{toshniwal2024openmathinstruct} augment extensive amount of math QA pairs from LLMs or humans as the additional training set to boost the performance.  Most recent studies apply a three-stage training paradigm to math-related corpus including pre-training, supervised fine-tuning, and reinforcement learning (e.g., InternLM2-Math \cite{ying2024internlm} and Deepseek-math \cite{shao2024deepseekmath}). 
%This paper diverges from previous research and benchmarks focused solely on improving LLMs' mathematical problem-solving capabilities. Instead, \mc~offers a thorough evaluation of broader competencies beyond single-round math question answering.

%However, the superior performance of these math-specialized LLMs comes with a cost, as shown in Figure \ref{fig:intro2}, although the SOTA models achieve similar accuracy with GPT3.5 and outperform Mistral significantly on GSM8K, these models lag behind GPT-3.5 and even Mistral-7B in following diverse math instructions and engaging in math dialogues on our \mc~benchmark. 
%Therefore, this paper diverges from previous research and benchmarks focused solely on improving LLMs' mathematical problem-solving capabilities. Instead, \mc~offers a thorough evaluation of broader competencies beyond single-round math question answering.

\paragraph{Multi-Turn Dialogues.}
The advancement of dialogue capabilities in LLMs, particularly their proficiency in multi-turn interactions, has been a key focus in LLM research \cite{ding2023enhancing,tunstall2023zephyr,zheng2024judging}. Collections of chat-style datasets, which include both human-to-LLM interactions (e.g., RealChat \cite{zheng2023realchat}) and LLM-to-LLM exchanges (e.g., Baize \cite{xu2023baize}, Ultrachat \cite{ding2023enhancing}), have been fundamental and instrumental in this area. Various benchmarks, such as MT-bench \cite{zheng2024judging} and MT-bench 101 \cite{bai2024mt}, have been developed to assess these capabilities. Most benchmarks primarily concentrate on general dialogues, while some datasets (e.g., MINT \cite{wang2023mint}) are specifically designed to evaluate specialized skills such as tool usage and feedback usage in reasoning tasks. In the realm of multi-turn dialogue on mathematics, many studies have discovered models' capability in aiding problem-solving \cite{wu2023empirical}, education \cite{macina2023mathdial,zhang2024mathemyths}, but the community still lacks a comprehensive benchmark and a study on how to empower LLMs to perform math conversation. Our work distinguishes itself by examining an under-explored direction of open-ended multi-turn dialogues: the benchmarking and analysis of combined mathematical reasoning and instruction-following on LLMs.% We believe that this exploration could facilitate novel applications of LLMs.
\vspace{-0.15cm}
\section{Conclusion}
\vspace{-0.1cm}
This paper introduces the \mc~benchmark as a new evaluative framework for assessing the capabilities of large language models (LLMs) in mathematical problem-solving and open-ended QA within multi-turn dialogue contexts. We demonstrate that while existing math-specialized LLMs excel at single-turn question-answering tasks, they significantly struggle with more complex, open-ended tasks that require understanding and following multi-turn instructions. We also collect and release a fine-tuning dataset \mcs~with math-centered dialogue interactions. LLMs trained with \mcs~show marked improvements in handling complex tasks in \mc~that require higher levels of comprehension and adaptability. %We believe \mc~serves not only as a tool for better understanding the current capabilities of LLMs in mathematical reasoning but also acts as a catalyst for future advancements in AI applications such as chatbots and education. 
%By highlighting specific areas of weakness, it guides researchers towards necessary improvements in LLM training approaches and model design, paving the way for more robust and versatile AI systems.

\section*{Broader Impact}
The research presented in this paper significantly advances automated mathematical reasoning and instruction-following in multi-turn interactions. By introducing the \mc~benchmark and the \mcs~dataset, this work enhances the capabilities of large language models (LLMs) in complex, interactive problem-solving. These advancements can lead to more effective educational tools and intelligent tutoring systems, supporting dynamic, interactive learning experiences. The public release of \mc~and \mcs~promotes further innovation and ensures responsible use, contributing to the development of AI systems that excel in real-world, context-aware applications.
% \section{Limitations}
% [1] diversity of the source dataset for \mc~construction

\bibliographystyle{plain}
\bibliography{reference}

\newpage
\appendix
\section{Appendix}

\subsection{Overall Results}
\label{app:overall}
To facilitate a more thorough and direct comparison across different models on our \mc~benchmark, we have formulated three comprehensive metrics based on two key aspects: problem-solving accuracy and open-ended task quality. Initially, we normalize all sub-metrics to a 0-1 scale. For problem-solving tasks, including follow-up QA and error correction, accuracies are normalized by dividing each by 100. For open-ended tasks, which are graded on a 1 to 5 scale, we normalize by dividing the scores by 5. We then define three metrics: 1) Overall Average: the average score of all ten sub-metrics listed in Tables \ref{tab:baseline} and \ref{tab:sft}; 2) Task Average: the average score across the four tasks; 3) Category Average: the average score of the two categories, i.e., problem-solving and open-ended QA.

The results in Table \ref{tab:overall}, based on the metrics defined above, indicate that the model with a Mistral backbone, fine-tuned with our \mcs~dataset, achieves the best performance across all three metrics. This proves the effectiveness of our SFT dataset and suggests that there is still potential for improvement in math-specific LLMs.

\begin{table}[H]
\footnotesize
\centering
\caption{Overall results of 7B LLMs. The best models are \textbf{bold} and the second best is underlined.}

\setlength\tabcolsep{3.5pt}
\renewcommand\arraystretch{1.1}
\begin{tabular}{l|c|c|c}
\hline
\textbf{Model} & \textbf{Overall Average} & \textbf{Task Average} & \textbf{Category Average} \\
\hline
LLaMA2-chat & 0.424 & 0.418 & 0.384 \\
Mistral-Instruct & 0.550 & 0.544 & 0.507 \\
Gemma-it & 0.463 & 0.463 & 0.432 \\
MAmmoTH & 0.422 & 0.442 & 0.424 \\
MetaMath & 0.451 & 0.470 & 0.463 \\
WizardMath & 0.454 & 0.492 & 0.476 \\
DeepSeek-Math & 0.452 & 0.500 & 0.476 \\
InternLM2-Math & 0.617 & \underline{0.635} & \underline{0.608} \\
Gemma-Math & 0.464 & 0.491 & 0.463 \\
Gemma-Math-IT & 0.508 & 0.528 & 0.511 \\
Gemma-Math-IT-Chat & 0.549 & 0.564 & 0.548 \\
Mistral-Math & 0.519 & 0.549 & 0.514 \\
Mistral-Math-IT & 0.565 & 0.586 & 0.557 \\
Mistral-Math-IT-Chat & 0.574 & 0.593 & 0.565 \\
Gemma-MathChat (Ours) & \underline{0.623} & 0.622 & \underline{0.608} \\
Mistral-MathChat (Ours) & \textbf{0.661} & \textbf{0.664} & \textbf{0.638} \\
\hline
\end{tabular}
\label{tab:overall}
\end{table}

\subsection{Experiment Details}
\subsubsection{Existing LLM Baselines}
\label{app:baseline}
% In our initial evaluation, 
We test three general-purpose, open-source models: LLaMA2-7B-chat \cite{touvron2023llama2}, Mistral-7B-Instruct \cite{jiang2023mistral} and Gemma-7B-it \cite{team2024gemma}. Additionally, we examine five math-specific LLMs: MAmmoTH \cite{yue2023mammoth} create and release MathInstruct, a math problem-solving dataset including CoT-style and PoT-style annotations and perform Supervised Fine-Tuning (SFT) on various base LLMs. In this paper, we use their released MAmmoTH-Mistral-7B variant. MetaMath-Mistral-7B \cite{yu2023metamath} is trained on augmented math data based on GSM8K and MATH. WizardMath-7B-v1.1~\cite{luo2023wizardmath} utilizes both SFT and reinforcement learning from evol-instruct Feedback on math instructions. InternLM2-7B-Math \cite{ying2024internlm} and DeepSeek-7B-Math \cite{shao2024deepseekmath} incorporate pre-training, SFT, and preference alignment focused on a mathematical corpus. We also present the performance of GPT-3.5-turbo, GPT-4-turbo and the latest GPT-4o. %We do not report on GPT-4 due to the evaluation tasks being generated by GPT-4, suggesting an inherent advantage for this model. Also, GPT-4 may learn towards its own responses considering its training data and architecture.

\subsubsection{Supervised Fine-tuning Implementation}

We utilize Mistral-7B and Gemma-7B as our backbone models and conduct fine-tuning using Low-Rank Adaptation (LoRA) \cite{hu2021lora}, with the rank set to 8 and alpha to 16. In our training process, we do not employ any specific templates or prefixes for the QA pairs but utilize the default chat template of the base models for transforming dialogues. The implementation is based on Pytorch along with the DeepSpeed \cite{rasley2020deepspeed} Library, and the models are trained on 8 NVIDIA V100 GPUs, each with 32GB of memory. We opt for float-16 (FP16) precision to decrease memory demands and computational requirements. The fine-tuning is carried out over three epochs, with a batch size of 32 and a learning rate of 3e-5. The cumulative training time for integrating all three types of datasets amounts to approximately 72 hours, and the training time for SFT with Math + \mcs~is around 30 hours.

\subsection{Error Type Analysis}

\begin{wrapfigure}{r}{0.35\textwidth}
  \centering
  \includegraphics[width=0.35\textwidth]{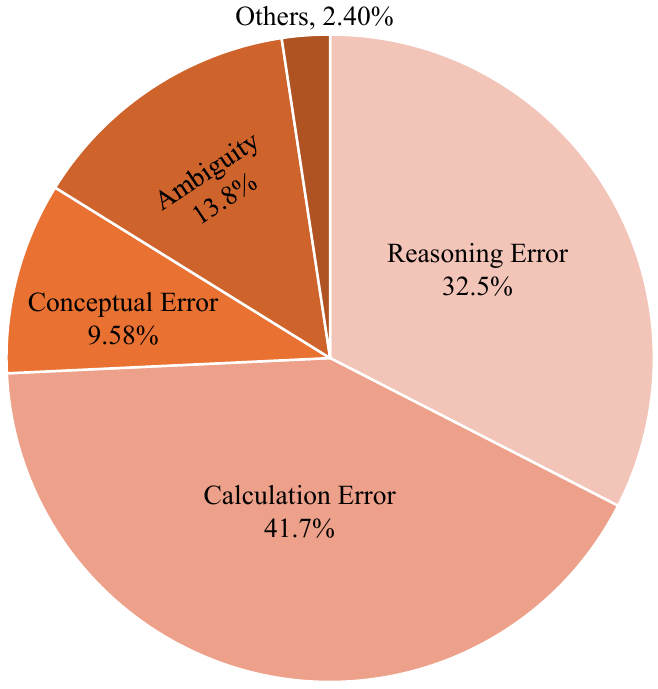}
  \caption{The distribution of error types on error correction task.}
  \label{fig:error}
\end{wrapfigure}

\label{app:error}
To ensure our benchmark contains a diverse array of error types, we randomly sampled 500 errors from our error correction task and used GPT-4 to determine their error types. The distribution of errors are shown in Figure \ref{fig:error}: Calculation Errors were most frequent, accounting for 41.8\% of the total. Reasoning Errors constituted 32.6\%, indicating challenges in logical thinking and strategizing the steps required to solve problems. Conceptual Errors, making up 9.6\%, pointed to difficulties in understanding underlying mathematical concepts. Ambiguity in solutions was noted in 13.8\% of cases, where the provided solution is ambiguous or unclear. This range of error types highlights the broad spectrum of challenges that MathChat contains, making our benchmark a robust tool for diagnosing and improving error correction and analysis ability across a variety of categories.

\subsection{Case Study}

\label{app:case}
\begin{figure}
  \centering
   \includegraphics[width=1.00\textwidth]{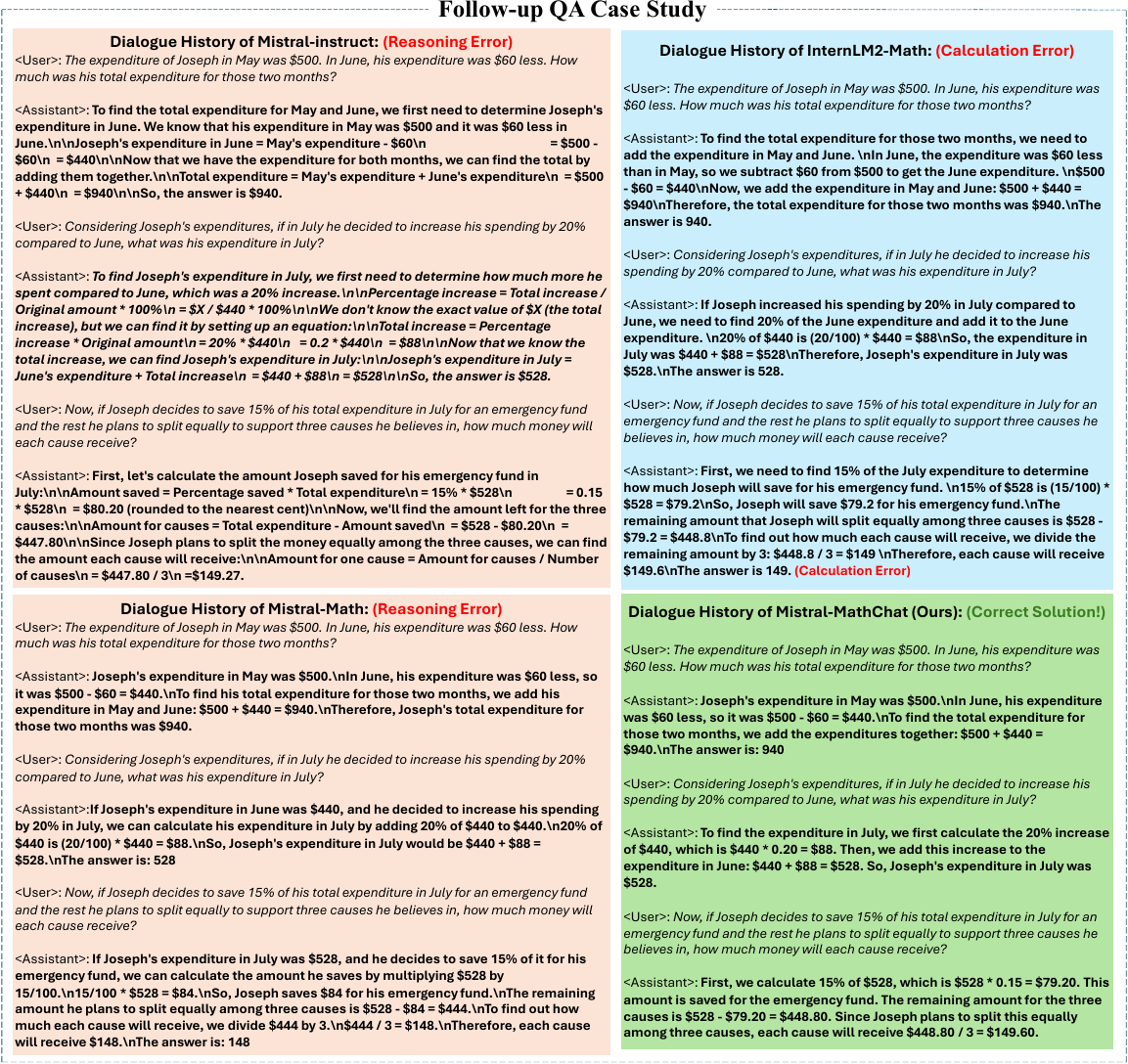}
  \caption{This figure illustrates the responses of four different LLMs in a follow-up QA task.}
  \label{fig:casefollowup}
\end{figure}

\paragraph{Follow-up QA} Figure \ref{fig:casefollowup} displays the responses from four LLMs on the follow-up QA task, specifically focusing on the third round of each model's response. The Mistral-instruct and Mistral-Math models, despite performing well in the first two rounds, exhibit reasoning errors in their third-round outputs. The InternLM2-Math model demonstrates a correct reasoning chain but makes a calculation error, resulting in an incorrect answer. These results indicate that the three models struggle with long-context reasoning, leading to increased errors as the number of dialogue turns rises. In contrast, our model, trained with \mcs, consistently performs well and successfully solves the third-round problem.

\begin{figure}
  \centering
   \includegraphics[width=1.00\textwidth]{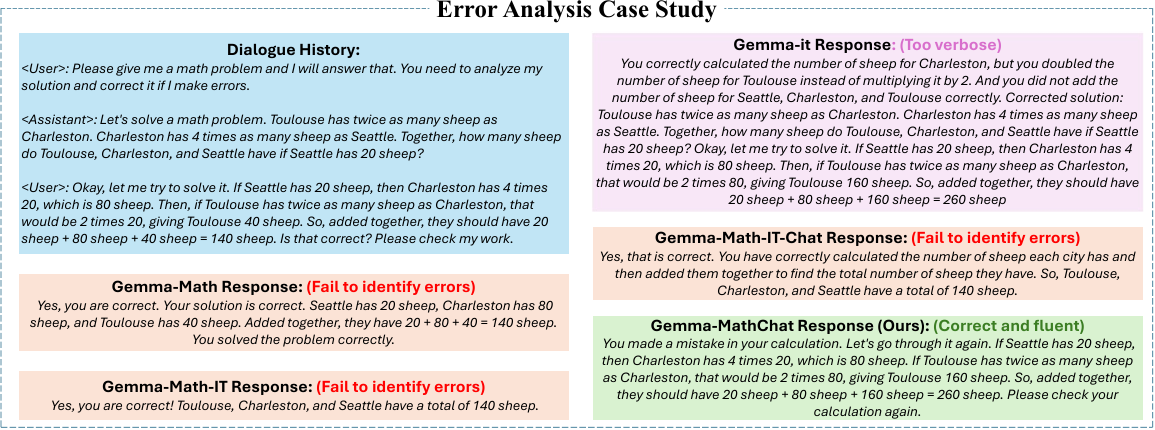}
  \caption{This figure illustrates the responses of four different LLMs in an error analysis task. The base Gemma-it model identifies and corrects the error but with verbosity, models fine-tuned with existing datasets fail to detect the error, and the \mcs~model successfully identifies and corrects the error succinctly and accurately.}
  \label{fig:casestudy}
\end{figure}

\paragraph{Error Analysis} Figure \ref{fig:casestudy} shows the responses from four LLMs on an error analysis task. This base model - Gemma-it 7B correctly identified the calculation error regarding the sheep in Toulouse and corrected the user's response. However, the feedback was verbose and included unnecessary details, potentially leading to confusion. The models fine-tuned with existing datasets failed to recognize the error in the user's solution, incorrectly affirming the erroneous calculation. This indicates a lack of training focus on error identification and correction capabilities in these models. Trained with our MathChat dataset, the model successfully identified and corrected the calculation error in a concise and clear manner. The analysis shows the importance of targeted, dialogue-rich training datasets like \mcs~in developing LLMs that are capable of effective educational interaction. The superior performance of the MathChat-trained model demonstrates its potential as a valuable tool in educational settings, offering precise and understandable corrections that can aid learning.

\begin{figure}
  \centering
   \includegraphics[width=1.00\textwidth]{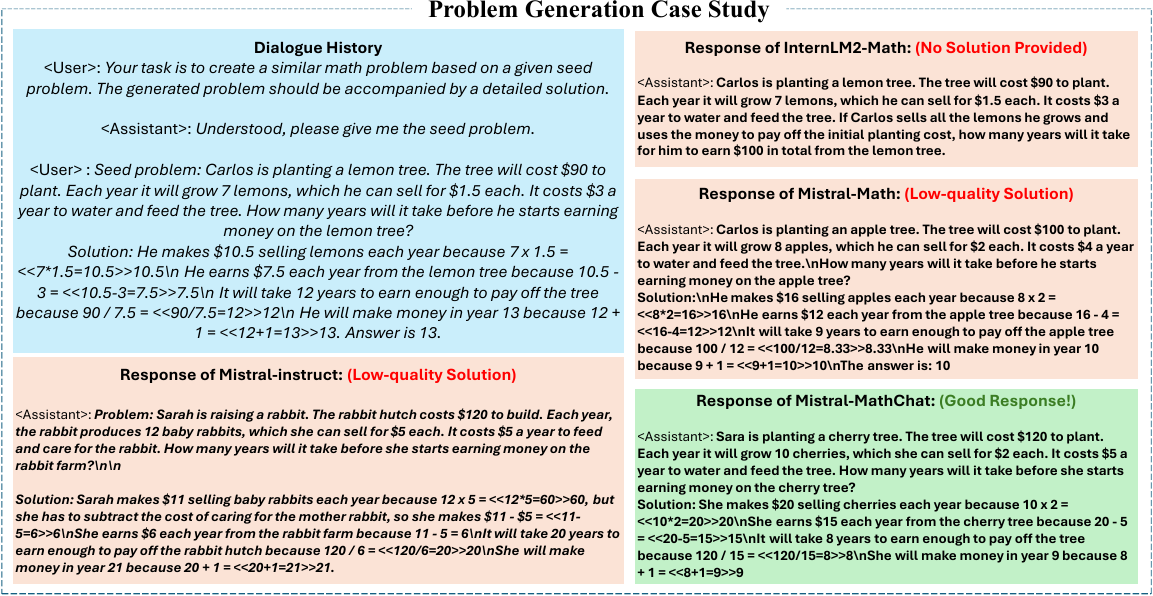}
  \caption{This figure illustrates the responses of four different LLMs in a problem generation task.}
  \label{fig:casep2p}
\end{figure}

\paragraph{Problem Generation} Figure \ref{fig:casep2p} presents the responses from four LLMs on the problem generation task. The Mistral-Instruct and Mistral-Math models provide incorrect solutions when generating a new problem, while the InternLM2-Math model fails to follow the instructions for generating solutions to the new problem. Only our model, trained with \mcs, successfully understands the instructions and delivers a high-quality solution.

\subsection{Prompt for Generating Task Data}
\begin{figure}[H]
  \centering
   \includegraphics[width=0.90\textwidth]{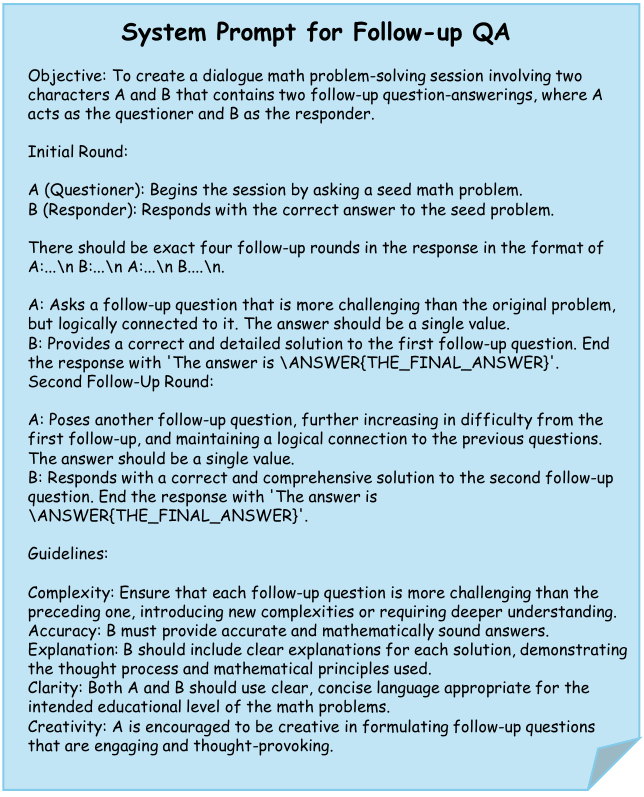}
  \caption{The system prompt for generating \textsc{Follow-up QA} task data.}
  \label{fig:prompt_followupqa}
\end{figure}

\begin{figure}[H]
  \centering
   \includegraphics[width=0.90\textwidth]{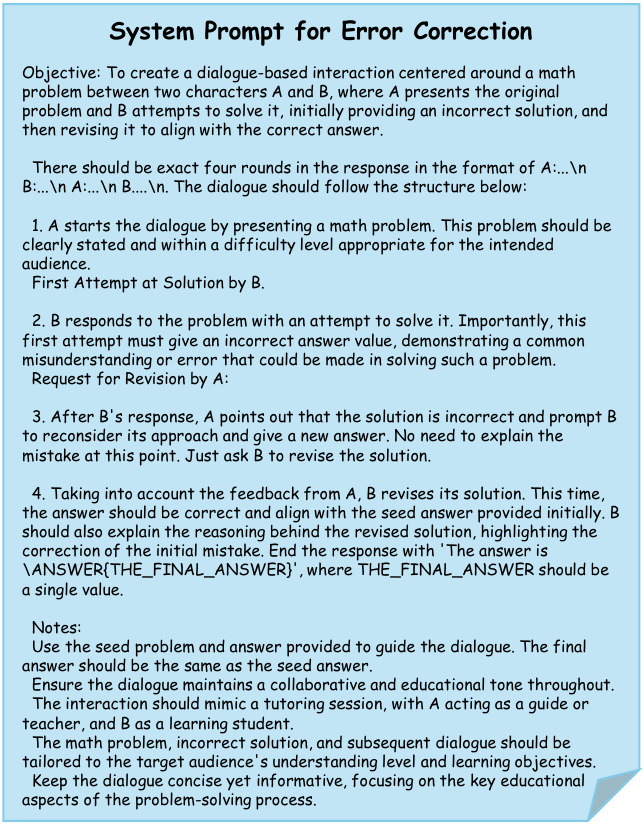}
  \caption{The system prompt for generating \textsc{Error Correction} task data.}
  \label{fig:prompt_errcor}
\end{figure}

\begin{figure}[H]
  \centering
   \includegraphics[width=0.90\textwidth]{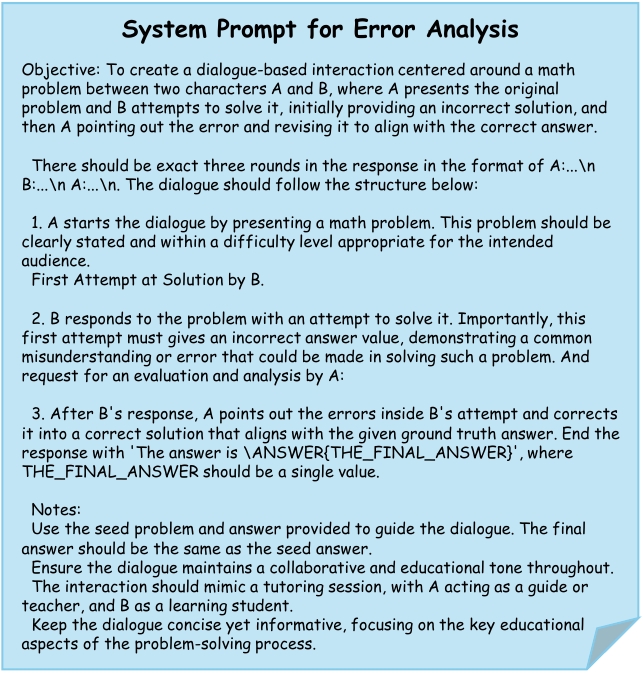}
  \caption{The system prompt for generating \textsc{Error Analysis} task data.}
  \label{fig:prompt_errana}
\end{figure}

\begin{figure}[H]
  \centering
   \includegraphics[width=0.90\textwidth]{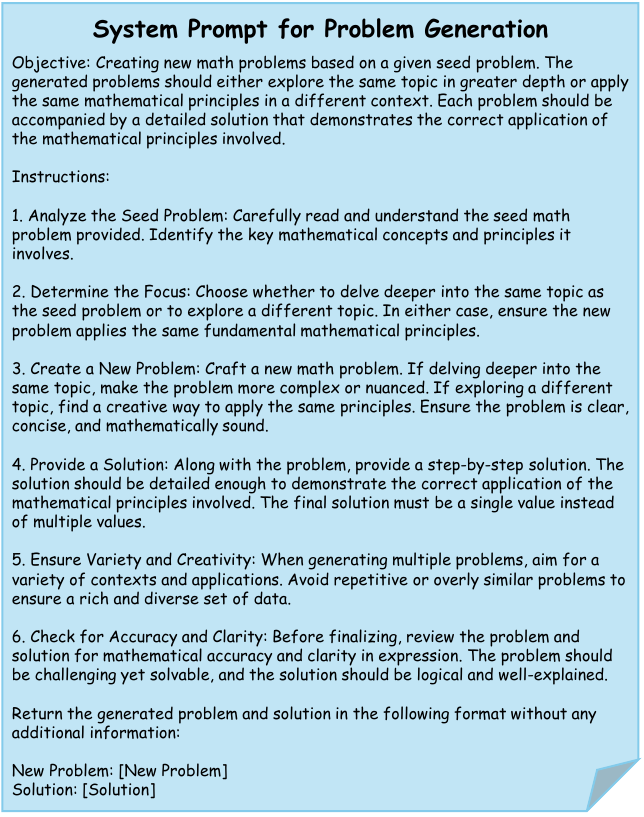}
  \caption{The system prompt for generating \textsc{Problem Generation} task data.}
  \label{fig:prompt_p2p}
\end{figure}

The prompts used to generate our MathChat benchmark is provided in Figure \ref{fig:prompt_followupqa}, \ref{fig:prompt_errcor}, \ref{fig:prompt_errana} and \ref{fig:prompt_p2p}.

\vfill
\vfill
\vfill
\vfill

\label{app:task_generation}

\newpage

\subsection{Prompt for Evaluating Open-ended Tasks}
\label{app:task_eval}

\begin{figure}[H]
  \centering
   \includegraphics[width=0.90\textwidth]{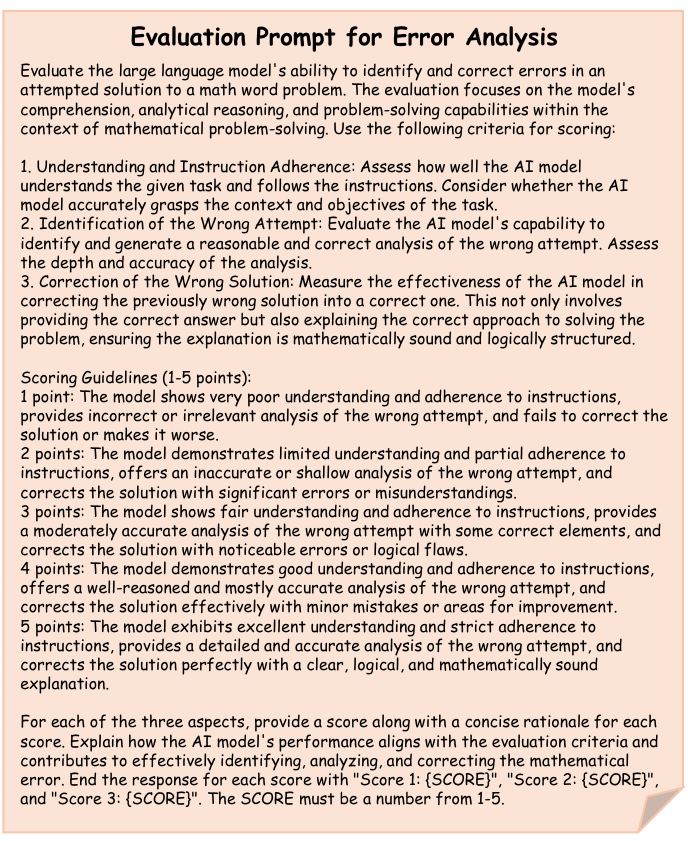}
  \caption{The system prompt for evaluating \textsc{Error Analysis} results using GPT-4.}
  \label{fig:eval_errana}
\end{figure}

\begin{figure}[H]
  \centering
   \includegraphics[width=0.90\textwidth]{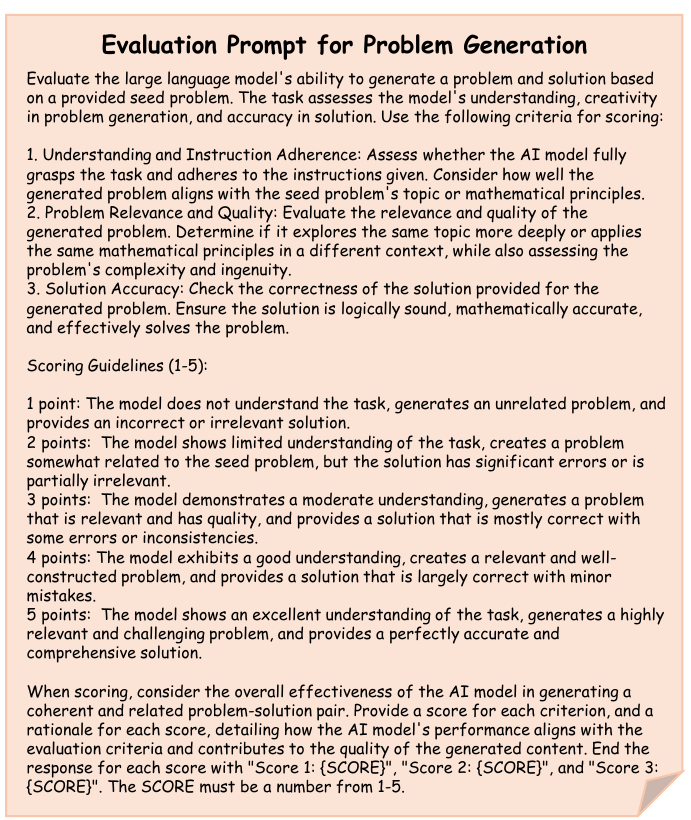}
  \caption{The system prompt for evaluating \textsc{Problem Generation} results using GPT-4.}
  \label{fig:eval_p2p}
\end{figure}

We provide the prompts used for evaluating the results of Error Analysis and Problem Generation in Figure \ref{fig:eval_errana} and \ref{fig:eval_p2p}.

\vfill
\vfill
\vfill
\vfill

\newpage

\subsection*{Prompt for Generating \mcs}
\begin{figure}[H]
  \centering
   \includegraphics[width=0.90\textwidth]{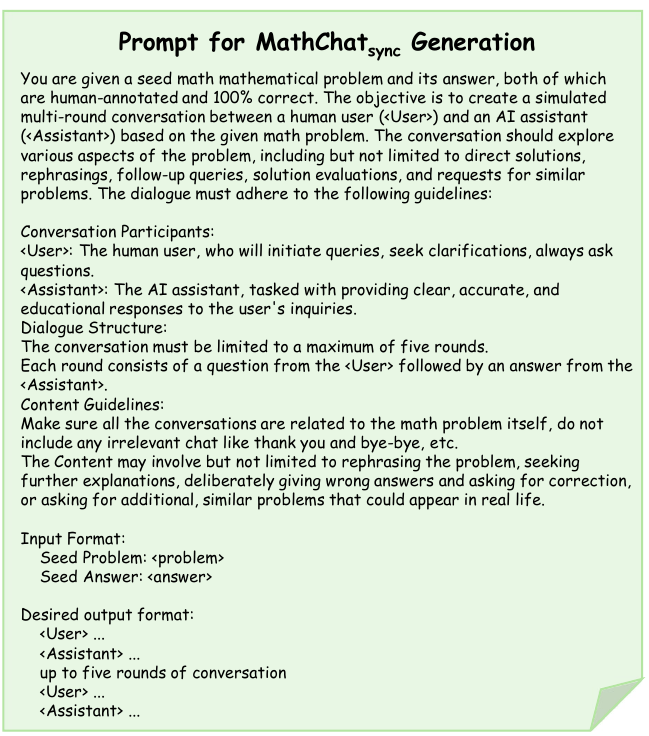}
  \caption{The system prompt for generating the \mcs~dataset for supervised fine-tuning.}
  \label{fig:prompt_mathchat}
\end{figure}

The prompt for generating \mcs~is shown in Figure \ref{fig:prompt_mathchat}.

\label{app:mathchat_generation}
% \subsection*{Examples of Chat Template of Specific LLMs}
% TODO
\end{document}